\documentclass{article}

\usepackage{PRIMEarxiv}

\usepackage[utf8]{inputenc} 
\usepackage[T1]{fontenc}    
\usepackage{hyperref}       
\usepackage{url}            
\usepackage{booktabs}       
\usepackage{amsfonts}       
\usepackage{nicefrac}       
\usepackage{microtype}      
\usepackage{lipsum}
\usepackage{fancyhdr}       
\usepackage{graphicx}       
\graphicspath{{media/}}     
\usepackage{amsmath,bm}

\title{Towards Real-Time Neural Video Codec for Cross-Platform Application Using Calibration Information
\thanks{\textit{\underline{Citation}}: 
\textbf{Kuan Tian, Yonghang Guan, Jinxi Xiang, Jun Zhang, Xiao Han, and Wei Yang. 2023. Towards Real-Time Neural Video Codec for Cross-Platform Application Using Calibration Information. In Proceedings of the 31st ACM International Conference on Multimedia (MM ’23), October 29–November 3, 2023, Ottawa, ON, Canada. ACM, New York, NY, USA, 10 pages. DOI:10.1145/3581783.3611955.}} 
}

\author{
  Kuan Tian \thanks{Equal contribution.} \\
  Tencent AI Lab \\
  Shenzhen, China \\
  \texttt{kuantian@tencent.com}
   \And
  Yonghang Guan \footnotemark[2] \\
  Tencent AI Lab \\
  Shenzhen, China \\
  \texttt{yohnguan@tencent.com}
   \And
  Jinxi Xiang \footnotemark[2] \\
  Tencent AI Lab \\
  Shenzhen, China \\
  \texttt{jinxixiang@tencent.com}
   \And
  Jun Zhang \thanks{Corresponding authors.} \\
  Tencent AI Lab \\
  Shenzhen, China \\
  \texttt{junejzhang@tencent.com}
   \And
  Xiao Han \\
  Tencent AI Lab \\
  Shenzhen, China \\
  \texttt{haroldhan@tencent.com}
   \And
  Wei Yang \\
  Tencent AI Lab \\
  Shenzhen, China \\
  \texttt{willyang@tencent.com}
}

\begin{document}
\maketitle

\begin{abstract}
The state-of-the-art neural video codecs have outperformed the most sophisticated traditional codecs in terms of rate-distortion (RD) performance in certain cases. However, utilizing them for practical applications is still challenging for two major reasons. 1) Cross-platform computational errors resulting from floating point operations can lead to inaccurate decoding of the bitstream. 2) The high computational complexity of the encoding and decoding process poses a challenge in achieving real-time performance. In this paper, we propose a \emph{real-time} \emph{cross-platform} neural video codec, which is capable of efficiently decoding ($\approx$25FPS) of 720P video bitstream from other encoding platforms on a consumer-grade GPU (e.g., NVIDIA RTX 2080). First, to solve the problem of inconsistency of codec caused by the uncertainty of floating point calculations across platforms, we design a calibration transmitting system to guarantee the consistent quantization of entropy parameters between the encoding and decoding stages. The parameters that may have transboundary quantization between encoding and decoding are identified in the encoding stage, and their coordinates will be delivered by auxiliary transmitted bitstream. By doing so, these inconsistent parameters can be processed properly in the decoding stage. Furthermore, to reduce the bitrate of the auxiliary bitstream, we rectify the distribution of entropy parameters using a piecewise Gaussian constraint. Second, to match the computational limitations on the decoding side for real-time video codec, we design a lightweight model. A series of efficiency techniques, such as model pruning, motion downsampling, and arithmetic coding skipping, enable our model to achieve 25 FPS decoding speed on NVIDIA RTX 2080 GPU. Experimental results demonstrate that our model can achieve real-time decoding of 720P videos while encoding on another platform. Furthermore, the real-time model brings up to a maximum of 24.2\% BD-rate improvement from the perspective of PSNR with the anchor H.265 (medium).
\end{abstract}

\keywords{Neural video codec \and cross-platform \and real-time codec}

\section{Introduction}

\begin{figure}[t!]
  \centering
  \includegraphics[width=1.0\linewidth]{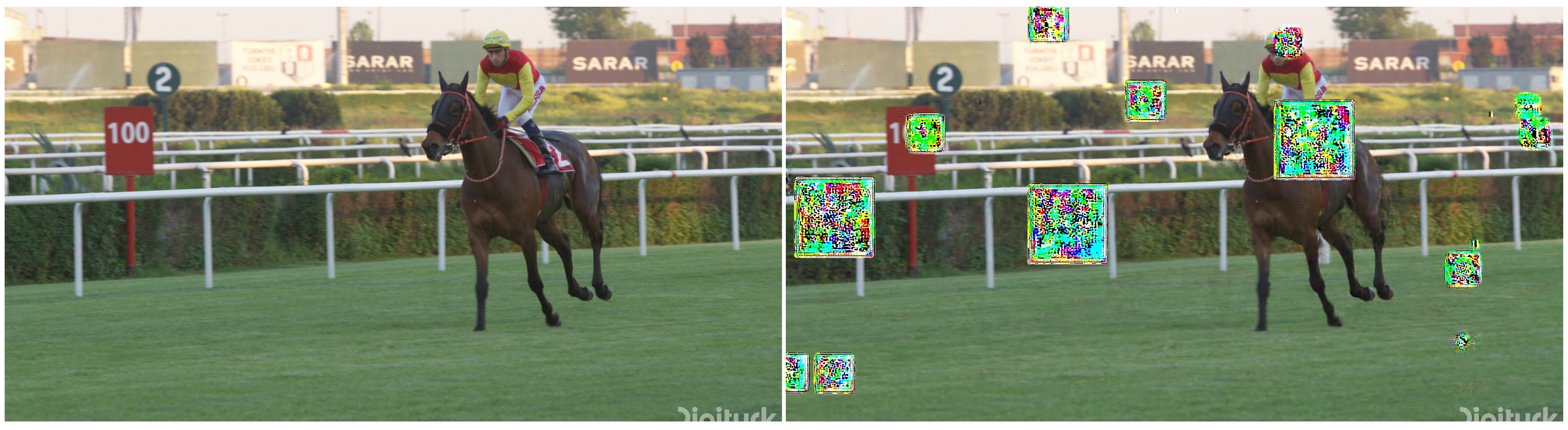}
  \caption{
  When the encoder and decoder run in cross-platform scenarios, the decoder will reconstruct an incorrect image on account of floating point math.
  }
  \label{fig:decode_error}
\end{figure}

In recent years, video codecs based on neural networks have attracted widespread attention and made significant progress in academic research. The latest neural video codecs (NVCs) have surpassed the state-of-the-art traditional video codecs (e.g., H.266/VTM) in terms of rate-distortion (RD) performance in certain cases \cite{bross2021overview,wang2023evc,li2023neural,li2022hybrid}. This will enable current and future high-definition videos to be stored and transmitted with less bitstream, benefiting almost all applications dealing with visual data \cite{lu2019dvc,agustsson2020scale,li2021deep,li2022hybrid,li2023neural,lin2020m,rippel2021elf,wang2023evc,zou2022devil}.

However, designing a real-time cross-platform NVC that can be applied in practice still faces two serious challenges. One is the cross-platform problem. In cross-platform scenarios, most learning-based video codecs face the issue of non-determinism \cite{balle2019integer}, such as incorrect reconstruction in Fig. \ref{fig:decode_error}. This is a common problem caused by floating point math on different hardware or software platforms, as numerical round-off is often platform dependent. Another challenge is to achieve real-time efficiency \cite{wang2023evc,liu2022convnet,chen2020variable}. With the improvement of NVC performance, the model complexity has significantly increased. The sophisticated model architecture and huge computational requirements make it impossible for NVCs to run in real-time on consumer-grade devices.

As initially defined by Balle et al. \cite{balle2019integer}, the non-determinism problem in cross-platform scenarios cannot be avoided when arithmetic coding is used for data compression \cite{balle2016end,balle2018variational,li2022hybrid,wang2023evc,li2023neural}. Existing methods mainly solve the non-determinism problem by using quantization techniques, which replace uncertain float calculations with deterministic integer calculations \cite{balle2019integer,sun2021learned,koyuncu2022device,he2022post}. Nevertheless, all these methods require more or less training steps for the model on calibration data, which makes it complicated to implement. In this paper, we propose a strategy that does not require any training and can \emph{partially} solve the inconsistency issue in cross-platform scenarios by transmitting a small amount of calibration information. Our method can maintain identifications between the training and inference stages and is easy to deploy in NVC algorithms. Specifically, we can achieve consistency between the encoder and decoder by encoding the entropy parameter coordinates that may cause errors in cross-platform computation into the transmitted bitstream. In addition, we propose to use a piecewise Gaussian to constrain the output of the entropy model, which effectively reduces the number of transmitted calibration coordinates and mitigates the impact of calibration transmitting on the model performance.

For real-time encoding and decoding, we use strategies including model pruning, motion downsampling, and arithmetic coding skipping to increase the decoding frame rate from 2 frames per second (FPS) to 25 FPS. Through model pruning \cite{he2017channel,li2016pruning,luo2020neural,he2019filter,molchanov2019importance,hinton2015distilling,zhao2022decoupled,wang2021distilling}, we reduce the computation amount of P-frame models from 1,100G Multiply-and-Accumulate operations (MACs) to 162G MACs and the computation amount of I-frame models from 950G MACs to 253G MACs for real-time computing with limited computing capacity. For the motion compensation module \cite{dosovitskiy2015flownet,ranjan2017optical,sun2018pwc,hui2018liteflownet,hui2020lightweight}, we downsample the input image from the original size to half to reduce the computation amount of the motion estimation model by four times. Due to the scale-invariant model architecture, only fine-tuning in the final optimization stage is required without the need to retrain the motion model. For another time-consuming arithmetic coding module \cite{duda2009asymmetric,witten1987arithmetic,howard1994arithmetic,cui2021asymmetric,guo2021learning,li2022hybrid,shi2022alphavc}, we use the skipping strategy to skip high-probability estimated parameters of the entropy model, reducing 60-70\% of the entropy coding computation amount without losing model performance \cite{shi2022alphavc}. Through these strategies, our video codec can increase the decoding speed in NVIDIA RTX 2080 GPU to 25 FPS, with the performance still surpassing the traditional H.265 (medium).

In conclusion, our method is capable of efficiently decoding 720P video bitstream in 25 FPS from other encoding platforms on a consumer-grade GPU, which still outperforms the traditional H.265 (medium). Our contributions are summarized as follows: 

\noindent\scalebox{1.5}{\textbullet}\ For the non-determinism issue caused by floating point math on different platforms, we propose a calibration information transmitting (CIT) strategy, which encodes the error-prone entropy parameter coordinates into the auxiliary bitstream, to achieve consistency between encoder and decoder. This allows us to achieve cross-platform compatibility without any additional training.

\noindent\scalebox{1.5}{\textbullet}\ To reduce the number of transmitted calibration coordinates, we propose a piecewise Gaussian constraint (PGC) that minimizes the number of output values of the entropy model that fall into a small range near the integer boundaries, thereby reducing the bitstream required for calibration information transmitting.

\noindent\scalebox{1.5}{\textbullet}\ A series of acceleration techniques, such as model pruning, motion downsampling, and arithmetic coding skipping, enable our model to achieve performance exceeding H.265 with medium preset at a 25 FPS decoding speed on NVIDIA RTX 2080 GPU.

\section{Related Work}

\subsection{Non-Determinism Issue in Cross-Platform}

The computational inconsistency problem in cross-platform scenarios is first defined by Balle et al. \cite{balle2019integer}. The reasons why other insensitive methods cannot avoid this common problem are analyzed in the paper because most encoding and decoding algorithms use arithmetic coding for data compression. Consequently, they proposed an integer-arithmetic-only network designed for learning-based image compression, to avoid the floating point math in the cross-platform \cite{balle2019integer}.
Koyuncu et al. implemented quantization to entropy networks with Gaussian mixture entropy model (GMM) and context modeling, which is more complex \cite{koyuncu2022device}. And He et al. used post-training quantization (PTQ) to train an integer-arithmetic-only model, thereby achieving a general quantization technique for image compression \cite{he2022post}. Summarizing the existing methods, they are very similar in concept to general model quantization techniques, such as quantization-aware training (QAT, \cite{jacob2018quantization,esser2019learned,bhalgat2020lsq+,krishnamoorthi2018quantizing,sun2021learned}) and post-training quantization (PTQ, \cite{nagel2019data,nagel2020up,li2021brecq}).

\begin{figure*}[t!]
  \centering
  \includegraphics[width=\linewidth]{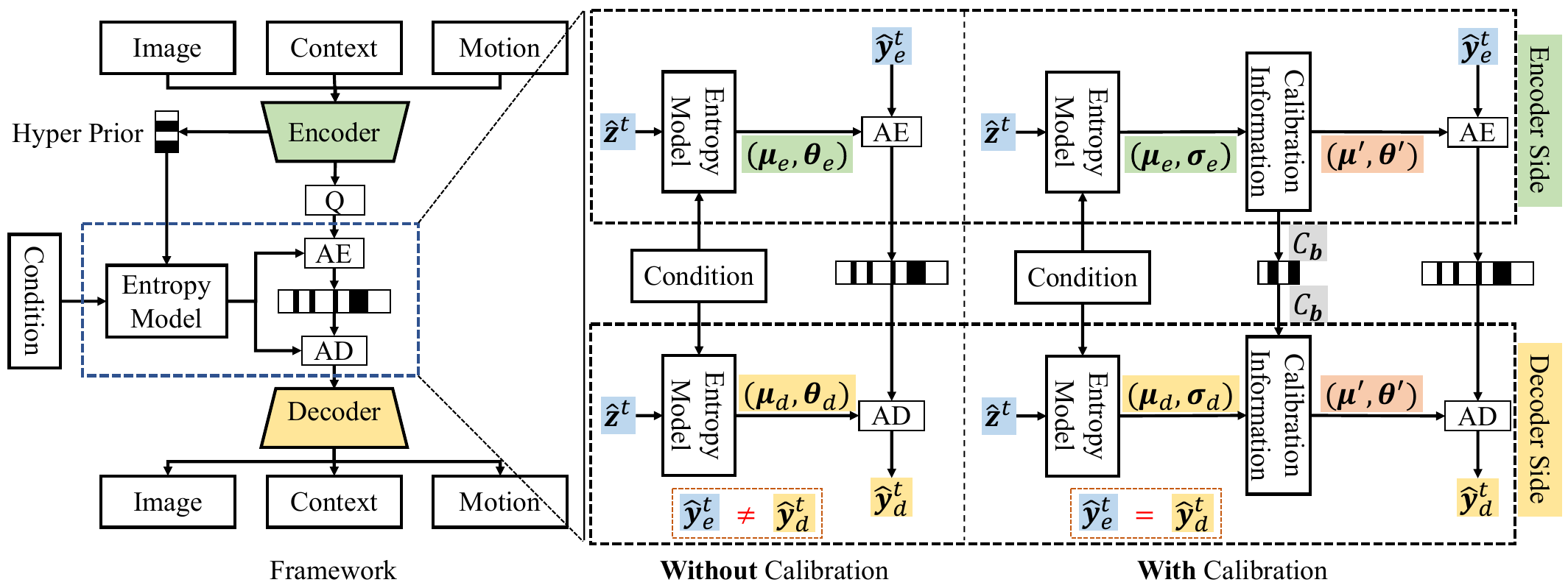}
  \caption{The framework of the proposed method addresses the computational inconsistency issue of video encoding/decoding in cross-platform scenarios through calibration information transmitting. \textit{Left}: We summarize the image and video encoding/decoding methods into the fundamental architecture, whose core is the entropy model. \textit{Middle}: In the absence of cross-platform strategies, decoding errors occur due to inconsistencies in floating-point calculations between the decoding and encoding sides. \textit{Right}: We design a calibration transmitting system to guarantee the consistent quantization of entropy parameters between the encoding and decoding stages. The parameters that may have transboundary quantization between encoding and decoding are identified in the encoding stage, and whose coordinates (i.e., $\textbf{C}_b$) will be delivered by auxiliary transmitted bitstream. Then, these inconsistent parameters can be processed properly in the decoding stage.}
  \label{fig:framework}
\end{figure*}

\subsection{Neural Video Compression}
Initially, Lu et al. developed the DVC model with all components in the traditional hybrid video codec replaced by an end-to-end neural network \cite{lu2019dvc}. Following this, DVC-Pro is proposed with more efficient residual/motion compression networks and corresponding refinement networks \cite{lu2020end}. To better handle disocclusions and fast motion failed cases, Agustsson et al. extended optical-flow-based estimation to a 3D transformation by proposing a scale-space flow \cite{agustsson2020scale}. Hu et al. compressed motion vectors using multi-resolution instead of single-resolution to optimize rate-distortion \cite{hu2020improving}. Yang et al. proposed a residual encoder and decoder based on RNN to exploit accumulated temporal information \cite{yang2020learning}.

Modified from the residual coding, DCVC employs contextual coding to compensate for the shortness of the residual coding scheme \cite{li2021deep}. While Mentzer et al. proposed simplifying the "hand-craft" video compression network of explicit motion estimation, warp, and residual coding with a transformer-based temporal model \cite{mentzer2022vct}. Contemporary work from Li et al. uses multiple modules, e.g., learnable quantization and parallel entropy model, to significantly improve the compression performance, which surpasses the latest VTM codec \cite{li2022hybrid}. AlphaVC introduces several techniques, e.g., conditional I-frame and pixel-to-feature motion prediction, to improve the rate-distortion performance \cite{shi2022alphavc}. There is also a work of real-time video decoding, called MobileCodec, which is the first-ever inter-frame neural video decoder running on a commercial mobile phone taking no account of the cross-platform issue \cite{le2022mobilecodec}.

Particularly, the entropy model is a crucial component of video compression and is mainly divided into non-autoregressive and autoregressive paradigms. The non-autoregressive methods only use temporal redundancy information from previous frames as priors when estimating entropy parameters \cite{lu2019dvc,agustsson2020scale,lu2020end,lin2020m}. While the autoregressive method uses both temporal redundancy information from previous frames, and spatial correlation within the current frame as priors to provide more information for the entropy parameters estimation \cite{li2021deep,minnen2018joint}. However, autoregressive prior is a serialized solution and follows a strict scanning order \cite{minnen2018joint}. Such a kind of solution is parallel-unfriendly which results in inferencing at a very slow speed. Then, several methods significantly reduce the number of autoregressive steps, which are much more time-efficient \cite{sheng2022temporal,li2023neural,xiangmimt,li2022hybrid,wang2023evc}. Furthermore, for arithmetic coding, almost all video compression models need to use it, resulting in an increase in runtime \cite{howard1994arithmetic}. Shi et al. proposed an efficient probability-based entropy parameter skipping strategy, which can significantly reduce the amount of calculation in arithmetic coding on both encoding side and decoding side \cite{shi2022alphavc}.

\section{Proposed Method}

The global framework follows a context conditioning entropy model proposed by Li et al. \cite{li2022hybrid}, which is summarized in Fig.~\ref{fig:framework}. The encoder is used to obtain the latents of the objectives such as $image$, $context$, and $motion$, while the decoder reconstructs the original information from the latents. The entropy model is designed to compress the latents $\boldsymbol{y}^t$ of input frame $\boldsymbol{x}^t$ effectively, where $\boldsymbol{x}^t$ is the frame at time step $t$. The accuracy of the entropy model's prediction (i.e., the estimated distribution of $\boldsymbol{y}^t$) determines the compression of $\boldsymbol{y}^t$ by arithmetic coding. The $condition$ is typically used to encode the context of inter-frames.
 
Specifically, the encoder is executed only at the encoding end, whereas the entropy model and decoder are executed both during the encoding and decoding stages. Consequently, slight variations in the entropy model computation may result in errors in the arithmetic coding and decoding, as discussed by Balle et al. \cite{balle2019integer}. Fig.~\ref{fig:decode_error} illustrates a decoding failed case that is attributed to errors that arise from non-deterministic floating point calculations. Different from the existing quantization-serials methods, we propose a new cross-platform video codec framework that can achieve cross-platform consistency without the need for training, simply by adding a small amount of calibration information in the transmitted bitstream.

\subsection{Preliminary}

The latest video compression model outperforms VTM due to the implementation of a robust entropy model, as evidenced by recent publications \cite{li2022hybrid,xiangmimt,mentzer2022vct}. These models accurately estimate the distribution of ${\boldsymbol{y}^t}$ from $\hat{\boldsymbol{z}}^t$ and $condition$ (if applicable). Assuming a normal distribution for $\boldsymbol{y}^t$, the estimated distribution of $\boldsymbol{y}^t$ can be derived as follows:
\begin{equation}
\begin{aligned}
    q\left(\boldsymbol{y}^t \mid \hat{\boldsymbol{z}}^t, \boldsymbol{c}\right)&=\prod_i\left(\mathcal{N}\left(\mu_i, \sigma_i^2\right) * \mathcal{U}\left(-\frac{1}{2}, \frac{1}{2}\right)\right)\left(y^t_i\right) \\
    \boldsymbol{\mu},\boldsymbol{\sigma}&=\mathrm{Etp}\left(\hat{\boldsymbol{z}}^t, \boldsymbol{c}\right),
\end{aligned}
\label{equ:ori_q_dist}
\end{equation}
where $i$ indicates the element index,  $\mathrm{Etp}$ is the entropy model, and $\boldsymbol{c}$ is an optional condition input. Since we subtract $\boldsymbol{\mu}$ from $\boldsymbol{y}^t$ before arithmetic coding, we use a zero mean distribution to obtain compressed bitstream.

To enhance the efficacy of encoding and decoding, we adopt the approach introduced by Balle et al. for nonlinearly discretizing $\boldsymbol{\sigma}$ into $L$ levels by introducing an intermediate variable $\ddot{\boldsymbol{I}}$ \cite{balle2019integer}, defined as follows:
\begin{equation}
\ddot{\boldsymbol{I}}=\mathrm{Q}\left(\boldsymbol{I}\right)=\left\lfloor \boldsymbol{I}\right\rfloor,
\label{equ:index_q}
\end{equation}
where 
\begin{equation}
\boldsymbol{I}=\mathrm{r}(\boldsymbol{\sigma})=\mathrm{clamp}\left(\frac{\log \left(\boldsymbol{\sigma}\right)-\log \left(\sigma_{min}\right)}{\sigma_{step}} ; 0, L-1\right),
\label{equ:index_ori}
\end{equation}
\begin{equation}
\sigma_{step}=\frac{\log \left(\sigma_{max}\right)-\log \left(\sigma_{min}\right)}{L-1}.
\label{equ:sigma_step}
\end{equation}
The parameters $\sigma_{max}$, $\sigma_{min}$, and $L$ determine the precision of the discretized $\boldsymbol{\sigma}$. $\mathrm{clamp}$ is a truncation function with a lower bound of 0 and an upper bound of $L-1$. In this context, the function $\mathrm{Q}$ represents the quantization process, which utilizes truncated quantization. The index range of $\ddot{\boldsymbol{I}}$ is $[0, L-1]$. 
Consequently, it is possible to generate a discrete $\boldsymbol{\theta}$ for $\boldsymbol{\sigma}$ by constructing a lookup table (LUT) of $\ddot{\boldsymbol{I}}$. Then, we express $\boldsymbol{\theta}$ as:
\begin{equation}
\boldsymbol{\theta}=\operatorname{LUT}\left(\ddot{\boldsymbol{I}}\right)=\exp \left(\log \left(\sigma_{min}\right)+\sigma_{step}\ddot{\boldsymbol{I}}\right).
\end{equation}

Therefore, the distribution used for $\boldsymbol{y}^t$ to arithmetic coding  in practical encoding and decoding is summarized as follows:
\begin{equation}
\begin{aligned}
    q_Q\left(\boldsymbol{y}^t \mid \hat{\boldsymbol{z}}^t,\boldsymbol{c}\right)&=\prod_i\left(\mathcal{N}\left(\mu_i, \theta_i^2\right) * \mathcal{U}\left(-\frac{1}{2}, \frac{1}{2}\right)\right)\left(y^t_i\right) \\
    \boldsymbol{\theta}&=\mathrm{LUT}\left(\mathrm{Q}\left(\mathrm{r}\left(\boldsymbol{\sigma}\right)\right)\right) \\
    \boldsymbol{\mu},\boldsymbol{\sigma}&=\mathrm{Etp}\left(\hat{\boldsymbol{z}}^t, \boldsymbol{c}\right).
\end{aligned}
\end{equation}

\subsection{Non-Determinism Issue}

The cross-platform failure in decoding is attributed to errors that arise from non-deterministic floating point calculations. Subsequently, we endeavor to explicate this matter in a comprehensible manner.
As depicted in the center of Fig. ~\ref{fig:framework}, each input frame $\boldsymbol{x}^t$ undergoes encoding to $\boldsymbol{y}^t$, followed by quantization to $\hat{\boldsymbol{y}}^t$. The hyperprior model transforms $\boldsymbol{y}^t$ into $\boldsymbol{z}^t$, which is then quantized to $\hat{\boldsymbol{z}}^t$. The entropy model at the encoder and decoder calculates the necessary entropy parameters $(\boldsymbol{\mu}, \boldsymbol{\theta})$ for arithmetic coding of the coincident $\hat{\boldsymbol{z}}^t$. 

However, due to the inconsistencies in floating-point calculations between the encoder and decoder, we obtain $(\boldsymbol{\mu}_e, \boldsymbol{\theta}_e)$ at the encoder and $(\boldsymbol{\mu}_d, \boldsymbol{\theta}_d)$ at the decoder. The encoder encodes $\hat{\boldsymbol{y}}^t_e$ into bitstream using $(\boldsymbol{\mu}_e, \boldsymbol{\theta}_e)$ through arithmetic coding, while the decoder decodes $\hat{\boldsymbol{y}}^t_d$ from the bitstream using $(\boldsymbol{\mu}_d, \boldsymbol{\theta}_d)$. Therefore, the decoder decompresses the incorrect $\hat{\boldsymbol{y}}^t_d$, resulting in the failure of the reconstructed image, as illustrated in Fig.~\ref{fig:decode_error}.

\subsection{Calibration Information}

\begin{figure*}[t]
  \centering
  \includegraphics[width=0.6\linewidth]{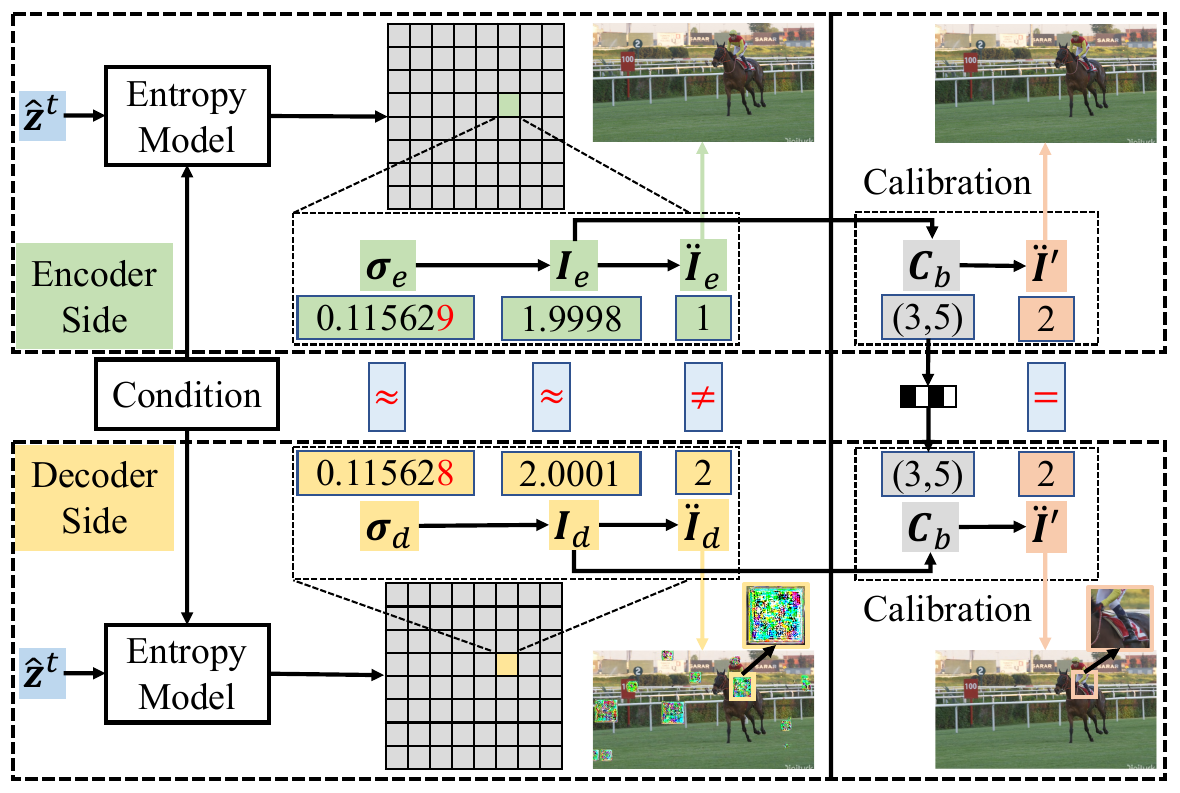}
  \caption{
   \textit{Left}: For the same input, due to the cross-platform operation of the encoder and decoder, the calculation of some values may result in inconsistent encoding and decoding, thereby causing the decoding end to obtain an incorrectly reconstructed image. \textit{Right}: By the proposed calibration transmitting system, we can align the probable inconsistent values of specific coordinates between the encoder and decoder, thus obtaining the correct reconstructed image.
  }
  \label{fig:calibration_detail}
\end{figure*}

 In the cross-platform scenario shown on the left side of Fig.~\ref{fig:calibration_detail}, mild inconsisitency in $(\boldsymbol{\mu}_e, \boldsymbol{\sigma}_e)$ and $(\boldsymbol{\mu}_d, \boldsymbol{\sigma}_d)$ caused by floating point calculations may result in completely different $\ddot{\boldsymbol{I}}$ according to Eq. ~\eqref{equ:index_q}~\eqref{equ:index_ori}~\eqref{equ:sigma_step}. To ensure consistency between the decoding and encoding ends, we propose a cross-platform calibration strategy, and whose specifics are elaborated in the right half of Fig.~\ref{fig:calibration_detail}.

Implicitly, $\boldsymbol{I}_e$ is derived from Eq. ~\eqref{equ:index_ori}~\eqref{equ:sigma_step}. To ensure seamless quantization across the encoding and decoding phases of $\boldsymbol{I}_e$, we employ a calibration precision $\epsilon$ and identify the distribution parameters (i.e., elements in $\boldsymbol{I}_e$) that may encounter transboundary quantization on the encoder side:
\begin{equation}
\mathcal{C}_b=\left\{(x, y, z)|\left(|\left(\mathrm{Q}\left(\boldsymbol{I}_e+\epsilon\right)-\mathrm{Q}\left(\boldsymbol{I}_e-\epsilon\right)\right) \mid>0\right)\right\},
\label{equ:calibration_boundary}
\end{equation}
where $\mathcal{C}_b$ is a set of coordinates $(x,y,z)$ in the range of $(C, H, W)$. We refer to $\mathcal{C}_b$ as the calibration information that is delivered via the auxiliary bitstream. In addition, we introduce another quantization function $\mathrm{Q}_D$ for $\boldsymbol{I}_e$ belonging to $\mathcal{C}_b$ to ensure consistency between the encoding and decoding ends, as:
\begin{equation}
\mathrm{Q}_D\left(\boldsymbol{I}\right)=\left\lfloor \boldsymbol{I}\right\rceil,
\end{equation}
where $\mathrm{Q}_D$ is a quantization function to round to the nearest integer value. Then determinate $\ddot{\boldsymbol{I}}^{\prime}$ can be obtained through follows: 

\begin{equation}
\ddot{\boldsymbol{I}}^{\prime}= \begin{cases}\mathrm{Q}\left(\boldsymbol{I}\right), & (\mathrm{x}, \mathrm{y}, \mathrm{z}) \notin \mathcal{C}_b \\ \mathrm{Q}_D\left(\boldsymbol{I}\right), & (\mathrm{x}, \mathrm{y}, \mathrm{z}) \in \mathcal{C}_b\end{cases}.
\label{equ:determinate_I}
\end{equation}

On the decoding side, we can obtain calibration information $\mathcal{C}_b$ from the transmitted bitstream. Then, the determinate $\ddot{\boldsymbol{I}}^{\prime}$ can be obtained through Eq. ~\eqref{equ:determinate_I} identically. Thus, we have completely consistent $(\boldsymbol{\mu}^{\prime}, \boldsymbol{\theta}^{\prime})$ in the decoding and encoding ends, as shown on the right side of Fig. ~\ref{fig:framework}. 
 This ensures that inconsistent distribution parameters can be properly processed during the decoding stage. Ultimately, the latent $\hat{\boldsymbol{y}}^t_d$ can be accurately restored from the compressed bitstream of $\hat{\boldsymbol{y}}^t_e$ using arithmetic coding, thus obtaining the correct reconstructed image.

\subsection{Calibration Bitstream Reduction}

In the context of video compression training, it is customary to impose constraints on the output of the entropy model $(\boldsymbol{\mu}, \boldsymbol{\sigma})$ through bit-rate loss:
\begin{equation}
R(p, q)=\mathbb{E}_{\boldsymbol{y}^t \sim p}\left[-\log _2 q\left(\boldsymbol{y}^t \mid \hat{\boldsymbol{z}}^t, \boldsymbol{c}\right)\right],
\label{equ:bit_rate}
\end{equation}
where $p$ is the true distribution of $\boldsymbol{y}^t$.  Hence, we proceed to address the rate-distortion optimization problem for video compression by formulating it as follows:
\begin{equation}
\mathcal{L}_{gen}=\lambda \cdot D+R,
\end{equation}
where $D$ represents the distortion between the original input frame $\boldsymbol{x}^t$ and the reconstructed frame $\hat{\boldsymbol{x}}^t$; $R$ denotes the bit-rate of all objectives to be compressed; $\lambda$ determines the trade-off between the number of bits and the distortion.

In practice, the values utilized for arithmetic coding are computed using Eq. ~\eqref{equ:index_q}~\eqref{equ:index_ori}~\eqref{equ:sigma_step}. 
However, when $\boldsymbol{I}$ is unconstrained during training, all values of $\boldsymbol{I}$ tend to randomly fall within the precision of $\epsilon$ as determined by Eq. ~\eqref{equ:calibration_boundary}, thereby increasing the length of the bitstream. 
To mitigate this issue and reduce the calibration bitstream, we propose a loss $\mathcal{L}_{c a l}$ based on piecewise Gaussian to push the transboundary values to the integer centers:
\begin{equation}
\mathcal{L}_{c a l}=\sum_{i=1}^{C \times H \times W} \mathrm{M}\left(\boldsymbol{I}_i\right) \cdot \left(\mathrm{G}(0.5, \delta)-\mathrm{G}\left(\boldsymbol{I}_i, \delta\right)\right)^2,
\label{equ:loss_cal}
\end{equation}
where
\begin{equation}
\mathrm{G}(x, \delta)=\frac{1}{\delta \sqrt{2 \pi}} e^{-\frac{1}{2}\left(\frac{{|}x-(\lceil x \rceil-0.5)|-0.5}{\delta}\right)^2},
\end{equation}

\begin{equation}
\mathrm{M}(x)= \begin{cases}0, & x<\eta \text { or }|x-(\lceil x \rceil-0.5)|<\eta \\ 1, & \text { others }\end{cases},
\end{equation}
$\delta$ is the standard deviation of the Gaussian function, $\eta$ is a threshold used to ignore certain ranges constrained by the loss function, $\lceil \cdot \rceil$ is a quantization function of ceiling, ($C, H, W$) represents the dimensions of the latent $\boldsymbol{y}^t$. Specifically, we only impose constraints on values whose distance to the interval center is greater than $\mathcal{\eta}$  because of the little possibility of transboundary for other values. Additionally, due to the presence of quantization in Eq. \eqref{equ:index_ori}, there is no possibility of transboundary for values close to integer 0, for which we do not need any constraints. 

By rectifying $\boldsymbol{I}$ during training with constraint $\mathcal{L}_{cal}$, $\boldsymbol{I}$ will have a more negligible probability of falling within the precision range, thereby reducing the total amount of $\mathcal{C}_b$ that needs to be transmitted and further decreasing the bitstream of the calibration system.

Through the implementation of a calibration strategy on the model trained with $\mathcal{L}_{gen}$, we can attain cross-platform consistency. Optionally, we can finetune the model with the following loss, which serves to minimize the bitstream required for transmitting the calibration information:
\begin{equation}
\mathcal{L} = \mathcal{L}_{gen} + \beta \cdot \mathcal{L}_{cal},
\end{equation}
where $\beta$ determines the strength of the piecewise Gaussian constraint $\mathcal{L}_{cal}$.

Furthermore, we denote $coord=\{0,1,2,...,C \times H \times W\}$ to represent the coordinate index of $\boldsymbol{I}$ after being flattened into a one-dimensional vector. Mathematically, we transform $\mathcal{C}_b$ into the coordinate index $\mathcal{V}_b$ according to:
\begin{equation}
\mathcal{V}_b=\left\{x \times H \times W + y \times W + z \mid(x, y, z) \in \mathcal{C}_b\right\},
\label{equ:vector_b}
\end{equation}
where $\mathcal{V}_b$ is a subset of $coord$. 

If $\mathcal{V}_b$ is transmitted directly, each coordinate index requires $log_2(C \times H \times W)$ bits to deliver. Considering that the coordinate indexes in $\mathcal{V}_b$ are monotonically increasing, we calculate relative coordinate index $\mathcal{V}_{rel}$ according to:
\begin{equation}
\mathcal{V}_{rel}(i)= \begin{cases} \mathcal{V}_b(i)- \mathcal{V}_b(i-1), & i>0 \\ \mathcal{V}_b(0), & i=0 \end{cases} ,
\label{equ:vector_rel}
\end{equation}
which is relative position coding (RPC). Therefore we can transmit each relative coordinate index with a bit length of $log_2(max(\mathcal{V}_{rel}))$.
This approach yields a substantial reduction in the overall bitstream $\mathcal{C}_b$ that needs to be transmitted, given that $max(\mathcal{V}_{rel}) \leq C \times H \times W$.

\subsection{Computing Efficiency Improvement}

\textit{\textbf{Model Pruning.}}
In order to enable real-time video decoding on user devices with limited computing capacity, we employ model pruning to reduce the model weight. Through this approach, we are able to significantly reduce the computational complexity of the I-frame model from its original 950G to 252G MACs, and that of the P-frame model from its original 1,100G to 162G MACs. This optimization ensures that the computational requirements of the model are well-suited to the computing capacity of the hardware.

\noindent\textit{\textbf{Motion Estimation Downsampling.}}
MobileCodec replaces the motion compensation by using a fully convolutional network to avoid warping \cite{le2022mobilecodec}. Unlike this, in order to reduce the computational complexity of motion estimation, we downsample the input image of the motion estimation network by a factor of 2 from the original size, which reduces the amount of calculation to 1/4 compared with the original setting. This can effectively save the computational requirements on the encoding side.

\noindent\textit{\textbf{Arithmetic Coding Efficiency.}}
In the conventional arithmetic coding process, each element of the feature $\boldsymbol{y}^t$ must be encoded to the bitstream using entropy parameters. However, we draw inspiration from the skipping strategy proposed by Shi et al. to reduce the number of elements that require calculation in arithmetic coding \cite{shi2022alphavc}, thereby significantly decreasing the coding time.

\begin{figure*}[h]
  \centering
  \includegraphics[width=1.0\linewidth]{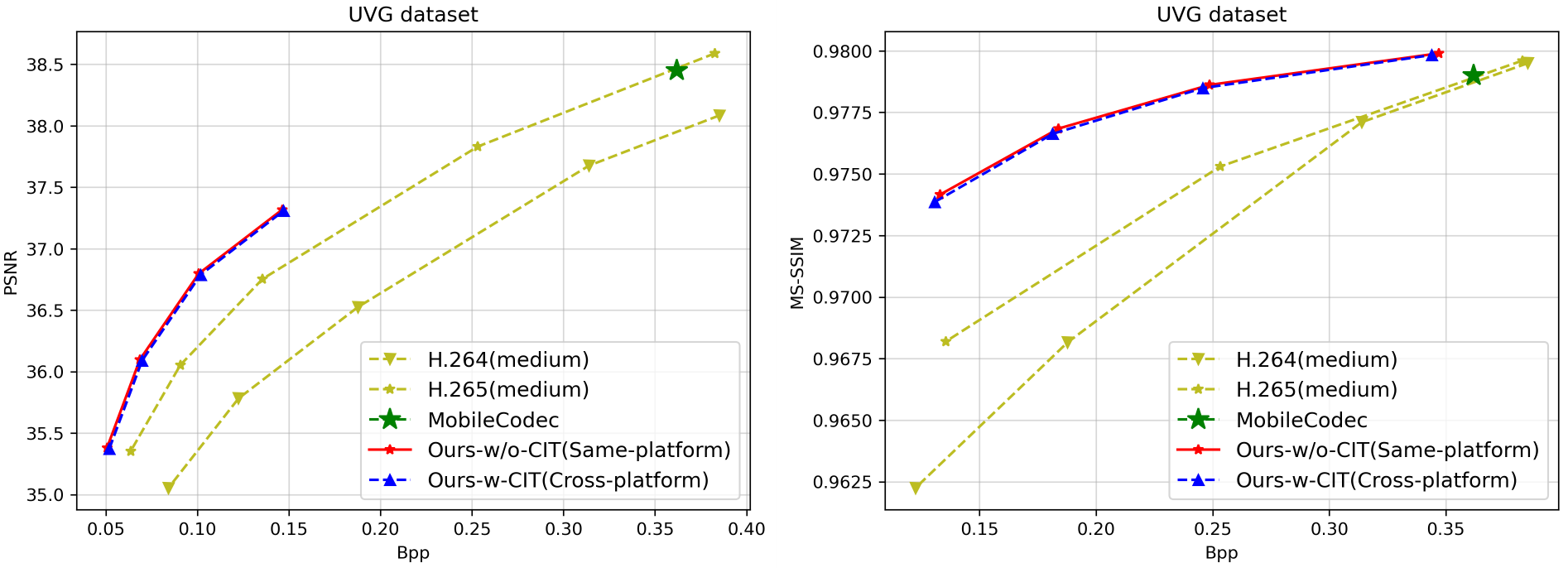}
  \caption{
  RD performance on UVG dataset.
  }
  \label{fig:rd_curve}
\end{figure*}

Specifically, for feature $\boldsymbol{y}^t$, whose quantized hyperprior is $\hat{\boldsymbol{z}}^t$. The distribution $(\boldsymbol{\mu}, \boldsymbol{\sigma})$ of $\boldsymbol{y}^t$ is predicted by the entropy model from $\hat{\boldsymbol{z}}^t$. For efficient computation, we use discretized $(\boldsymbol{\mu}, \boldsymbol{\theta})$ for arithmetic coding. When the model has high certainty about an element, i.e., when $\theta_i$ is very small, we can skip the arithmetic coding of element $y^t_i$ and use the estimated value $\mu_i$ from the model instead, where $i$ is an index in the feature map. This approach enables us to significantly reduce the number of elements for arithmetic coding, thereby reducing arithmetic coding time.

\section{Experiments}

\subsection{Experimental Setup}

\textit{\textbf{Datasets.}} 
Our model is trained on Vimeo-90k dataset \cite{xue2019video}, which contains 89800 video clips with the resolution of $448 \times 256$. Video frames are randomly cropped into 256x256 patches during training. We evaluate our model using the UVG dataset \cite{mercat2020uvg}, which has a resolution of $1920 \times 1080$. To compare model performance, we test each video for 96 frames with GOP size 12. We obtain $1920 \times 1024$ images by center cropping from the original resolution to ensure that the input image shape is divisible by 64. Additionally, to test cross-platform encoding and decoding speed, we crop 1080P images into $1280 \times 768$ (720P) images using the center crop window.

\noindent\textit{\textbf{Implementation and training details.}} 
The training process is divided into two main parts. 
In the first part, we pre-train the model using the rate-distortion loss $\mathcal{L}_{gen}$, following the previous related works \cite{guo2021learning,sheng2022temporal,li2023neural}. We begin by independently training the I-frame model and P-frame model. Then we jointly train the I and P frame models with the loss function $\mathcal{L}_{gen}$ and set the training group of pictures (GOP) to 7 (because each clip of Vimeo-90k contains 7 frames). This process enables us to obtain the optimal video compression model. 
In the second part, we finetune the compression model by adding calibration constraints $\mathcal{L}_{cal}$ to obtain the final deployment version. Specifically, since the calibration constraint only applies a slight rectification of the model's output, we freeze all modules except those related to the entropy model during finetuning.

As the method proposed by Li et al. \cite{li2022hybrid}, we set 4 $\lambda$ values to (85, 170, 380, 840) to optimize for PSNR-BPP, and (10, 20, 40, 100) for SSIM-BPP. And we discretize $\boldsymbol{\sigma}$ with $L$ set to 32, $\sigma_{min}$ set to 0.01, and $\sigma_{max}$ set to 64. To reduce the calibration bitstream, we set $\delta$ to 1.0, $\eta$ to 0.3, and $\beta$ to 1.0.

\begin{table}[t]
\caption{BD-rate calculated by \textit{PSNR} and \textit{SSIM} on UVG dataset with respect to the anchor H.265 (medium).}
\centering
\begin{tabular}{c|ccc}
\hline 
\hline 
& & Ours w/o CIT & Ours w CIT \\
& H.264 & same-platform & cross-platform \\
\hline
PSNR-BPP & 45.8 \% & -25.6 \% & -24.2 \% \\
\hline
SSIM-BPP & 32.6 \% & -34.1 \% & -33.7 \% \\
\hline
\end{tabular}
\label{tab:bd_rate}
\end{table}

\noindent\textit{\textbf{Cross-platform configuration.}} 
We test our model's efficiency for encoding and decoding across platforms using an \textit{NVIDIA Tesla A100} machine for the encoder and an \textit{NVIDIA RTX 2080} machine for the decoder, with \textit{AMD EPYC 7K62} and \textit{Intel Xeon W-2133} CPUs, respectively. We use the latest PyTorch 2.0 environment and perform FP32 inference on the models.

\subsection{Real-time Cross-platform Results}
We present the experimental results of the proposed method from two perspectives. 1) We demonstrate the overall performance of our method running cross-platform through the rate-distortion curves. 2) The computational efficiency is further evaluated for the I-frame model and the P-frame model, respectively.

\begin{table*}[t]
\caption{Decompression failure rates across different platforms. With calibration information transmitting, all decoding failures have been resolved.}
\centering
\begin{tabular}{lrrrrrrrr}
\hline 
\hline
compressed on & CPU 1 & CPU 1 & CPU 1 & CPU 1 & GPU 1 & GPU 1 & GPU 1 & GPU 1\\
decompressed on & CPU 1 & CPU 2 & GPU 1 & GPU 2 & CPU 1 & CPU 2 & GPU 1 & GPU 2\\
\hline
\multicolumn{9}{c}{Test Data: 96 frames of UVG ReadySteadyGo Video of 1920x1024 pixels} \\
\hline
w/o CIT & 0\% & 94.8\% & 99.0\% & 96.9\% & 96.9\% & 96.9\% & 0\% & 100\%\\
w   CIT($\epsilon=$1e-4) & 0\% & 0\% & 0\% & 0\% & 0\% & 0\% & 0\% & 0\%\\
w   CIT($\epsilon=$1e-3) & 0\% & 0\% & 0\% & 0\% & 0\% & 0\% & 0\% & 0\%\\
w   CIT($\epsilon=$1e-2) & 0\% & 0\% & 0\% & 0\% & 0\% & 0\% & 0\% & 0\%\\
\hline
  \multicolumn{5}{l}{CPU 1: AMD EPYC 7K62 48-Core Processor} &\multicolumn{4}{l}{GPU 1: NVIDIA Tesla A100} \\
  \multicolumn{5}{l}{CPU 2: Intel Xeon W-2133} &\multicolumn{4}{l}{GPU 2: NVIDIA RTX 2080} \\
\end{tabular}
\label{tab:ab_cross_platform}
\end{table*}

\noindent\textit{\textbf{Rate-Distortion Performance.}} 
We evaluate the performance of all models using common metrics such as PSNR and MS-SSIM \cite{wang2004image}. As shown in Fig.~\ref{fig:rd_curve}, the solid red line represents the performance of our model when encoding and decoding within the same platform without calibration information transmitting.

When decoding across platforms, decoding errors occur, making it impossible to draw a curve for the model's cross-platform decoding result. However, the blue dashed line indicates the metrics of our model when decoding on a different platform with calibration information transmitting (e.g., encoding on A100 and decoding on 2080). This result demonstrates that after adopting the calibration strategy, our model can successfully decode all video frames in cross-platform scenarios.

Furthermore, we compare our model's performance with MobileCodec \cite{le2022mobilecodec}, which is the first-ever inter-frame neural video decoder running on a commercial mobile phone but does not account for the cross-platform issue. The green star in the figure represents the result of MobileCodec.

We also evaluate the compression performance using BD-rate \cite{bjontegaard2001calculation} computed from PSNR-BPP and SSIM-BPP, respectively. The conventional codec H.265 in FFmpeg is used as the anchor with medium preset. As shown in Table. \ref{tab:bd_rate}, when the decoding end runs on the same platform, our model achieves a 25.6\% bitrate saving. From the perspective of SSIM, the improvement is even larger, with a 34.1\% bitrate saving.
In cross-platform scenarios, using the proposed method enables successful decoding of all video frames, with only a 1.4\% BD-rate decrease on PSNR-BPP and a 0.4\% BD-rate decrease on SSIM-BPP, respectively.

\noindent\textit{\textbf{Efficiency.}}
Considering that the I-frame model is simpler than the P-frame model, we design the I-frame model to contain more parameters and larger computational complexity in our lightweight model. However, due to its simple architecture, the actual computation time of the I-frame model is less than that of the P-frame model. Detailly, as shown in Table. \ref{tab:params_and_macs}, in our cross-platform test scenario, we can achieve an average encoding time of 35.6 ms and 28.1 FPS on the A100 machine, and an average decoding time of 39.7 ms and 25.2 FPS on the 2080 machine, respectively. We calculate the average encoding and decoding time per frame in units of GOP, with each GOP consisting of 1 I frame and 11 P frames, totaling 12 frames. We have recorded a demo video of real-time cross-platform decoding on a 2080 GPU, which will be shown in the Supplementary Materials.

\begin{table}[t]
\caption{Model complexity and computation speed. The encoder runs on an A100 machine, and the decoder runs on a 2080 machine.}
\centering
\begin{tabular}{c|rr|c}
\hline
\hline Module & Params & MACs & Encoding time(A100, FP32) \\
\hline
I model & 23.71 M & 252.633 G & 26.0 ms \\
P model & 10.17 M & 161.686 G & 36.5 ms \\
\hline
\hline Module & Params & MACs & Decoding time(2080, FP32) \\
\hline
I decoder & 20.95 M & 160.683 G & 37.1 ms \\
P decoder &  8.75 M &  76.311 G & 39.9 ms \\
\hline
\end{tabular}

\label{tab:params_and_macs}
\end{table}

As our current tests are all running on PyTorch 2.0 without parallel acceleration, there is still much room for engineering implementation, parallel optimization, and other work that can be done based on the proposed method. In the future, real-time decoding on consumer-grade GPUs of 1080P video is also highly feasible.

\subsection{Ablation Study}
We conduct a series of ablation studies to demonstrate the effectiveness of different components.

\noindent\textit{\textbf{Calibration Information Transmitting.}}
In order to address the cross-platform calculation inconsistency problem, we propose the calibration information transmitting method. To demonstrate the effectiveness, we evaluate the encoding and decoding on four different platforms (two CPU platforms and two GPU platforms), using the UVG dataset. As shown in Table. \ref{tab:ab_cross_platform}, when the cross-platform strategy is not used, the encoding and decoding between different platforms will result in a decoding failure of $94\%$-$100\%$ of video frames. In contrast, all four platforms achieve $0\%$ failure on the encoding and decoding platforms with the proposed calibration transmitting strategy, demonstrating the effectiveness of our method for addressing the calculation inconsistency problem.

Encountered with cross-platform issues, almost all general codec models will lead to decoding failures. Existing methods solve this problem through quantization-like methods. While we solve this issue without modifying the model, only by transmitting a small amount of calibration information. And our method theoretically works for all video compression algorithms that use similar arithmetic coding methods.

\begin{figure}[h]
  \centering
  \includegraphics[width=0.5\linewidth]{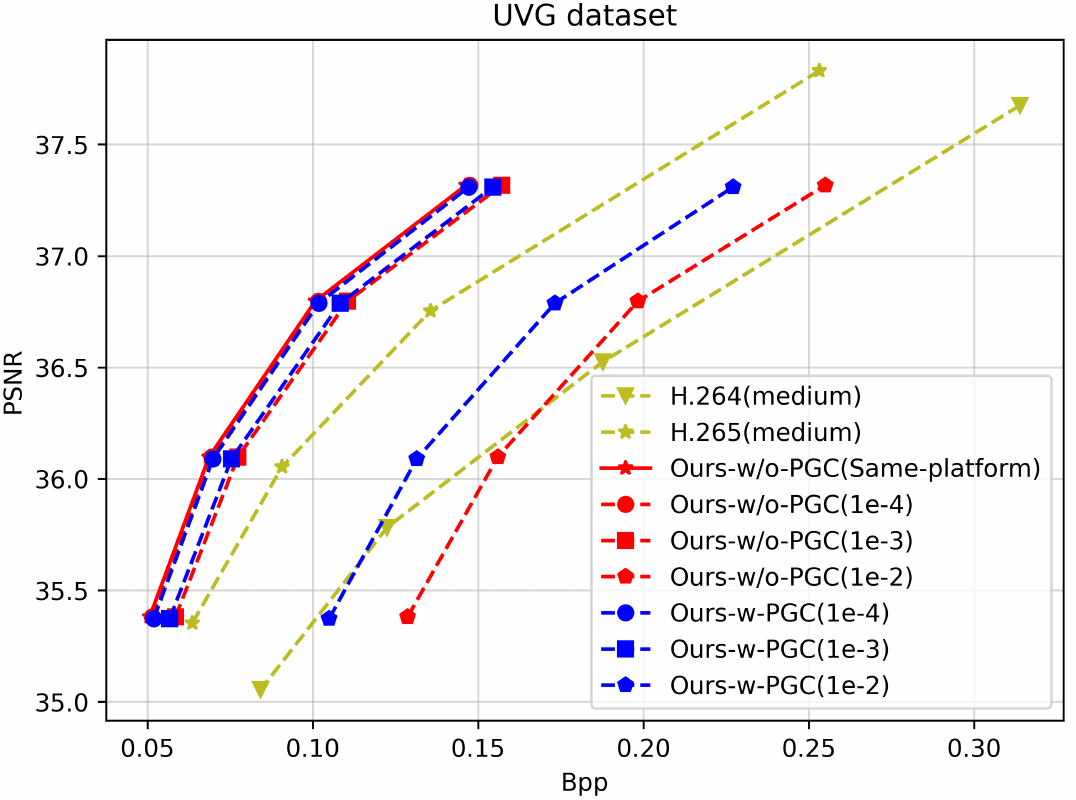}
  \caption{
  Ablations on the piecewise Gaussian constraint. When there are different error levels between the encoding and decoding ends, we will get different RD curves shown as dashed lines.
  }
  \label{fig:ab_calibration}
\end{figure}

\noindent\textit{\textbf{Piecewise Gaussian Constraint.}}
Table. \ref{tab:ab_cross_platform} shows that introducing a different level of precision of calibration information transmitting can effectively achieve consistency in cross-platform encoder and decoder. But higher errors between the encoding and decoding side will result in more calibration bitstream to be transmitted. We evaluate the UVG data and count the average amount of calibration transmitting required by the different errors between the encoder and decoder. As shown in Fig.~\ref{fig:ab_calibration}, with piecewise Gaussian constraint, our model can save a significant amount of bitstream of calibration information to be transmitted. To ensure a fair comparison, we transmit each value of different precision using the same bit length (i.e., 16 bits for every calibration coordinate).

\begin{table}[t]
\caption{Comparison of BD-rate and quantity of calibration transmitting using piecewise Gaussian constraint.}
\centering
\begin{tabular}{lrrr}
\hline
\hline
 Precision & w/o PGC & w PGC & Improved \\
\hline
\multicolumn{4}{c}{\textbf{BD-rate}} \\
\hline
$\epsilon=$0     & -25.6\% & -     & - \\
$\epsilon=$1e-4  & -24.8\% & -24.2\% & 0.6\% \\
$\epsilon=$1e-3  & -17.1\% & -18.7\% & \textbf{-1.6\%} \\
$\epsilon=$1e-2  &  59.4\% &  36.7\% & \textbf{-22.7\%} \\
\hline
\multicolumn{4}{c}{\textbf{Quantity of CIT}} \\
\hline
$\epsilon=$1e-4  & 114.8   & 82.4    & \textbf{-28.2\%} \\
$\epsilon=$1e-3  & 1141.7  & 825.7   & \textbf{-27.7\%} \\
$\epsilon=$1e-2  & 11414.9  & 8256.7 & \textbf{-27.7\%} \\
\hline
\end{tabular}
\label{tab:ab_cal_diff_epsilon}
\end{table}

\begin{table}[t]
\caption{Bits comparison of relative position coding.}
\centering
\begin{tabular}{r|rr|rr}
\hline 
\hline 
                   & w/o PGC & w/o PGC & w PGC & w PGC \\
Precision & w/o RPC & w RPC & w/o RPC & w RPC \\
\hline
$\epsilon=$1e-4 & 114.8$\times$ 20 & $\times$ 13 & 82.4 $\times$ 20 & $\times$ 14 \\
$\epsilon=$1e-3 & 1141.7 $\times$ 20 & $\times$ 10 & 825.7 $\times$ 20 & $\times$ 10 \\
$\epsilon=$1e-2 & 11414.9 $\times$ 20 & $\times$ 7 & 8256.7 $\times$ 20 & $\times$ 7 \\
\hline
Average & -0\%  & \textbf{-63.4\%} & \textbf{-27.7\%} & \textbf{-73.5\%} \\
\hline

\multicolumn{3}{r}{${log}_2 (737280)=20$} &
\multicolumn{2}{r}{${log}_2 (737280/83)=14$} \\
\multicolumn{3}{r}{${log}_2 (737280/115)=13$} &
\multicolumn{2}{r}{${log}_2 (737280/826)=10$} \\
\multicolumn{3}{r}{${log}_2 (737280/1142)=10$} &
\multicolumn{2}{r}{${log}_2 (737280/8257)=7$} \\
\multicolumn{5}{r}{${log}_2 (737280/11415)=7$} \\
\hline
\end{tabular}
\label{tab:bytes_compare}
\end{table}

As shown in Table. \ref{tab:ab_cal_diff_epsilon}, it illustrates that introducing piecewise Gaussian constraint for the output of the model can effectively reduce the amount of calibration information transmitting at the same error level, which brings us up to a maximum of 22.7\% BD-rate improvement from the perspective of PSNR. Slightly different in a small range, such as $\epsilon$ is the same as 1e-4, the BD-rate of introducing calibration transmitting strategy in the original model is even higher than that of the model finetuned with piecewise Gaussian constraint, which is caused by the larger subtle model deviation introduced by finetuning than the improvement of the piecewise Gaussian constraint. The advantage of rectified distribution cannot be illustrated with very few potential error coordinates. 

The second half of the Table. \ref{tab:ab_cal_diff_epsilon} particularly shows the average coordinate number of calibration transmitting required for different precision levels. It can be seen that the number of calibration coordinates is reduced by 27.9\% on average after using the constraint.

Furthermore, Table. \ref{tab:bytes_compare} shows that by applying relative position coding, the bits required for transmitting calibration information can be reduced by up to 73.5\% for different settings. It should be added that we assume each calibration information position is randomly distributed in the latents, so we can estimate the number of bits required for each calibration coordinate based on the average number of transmitted calibration information. Then, we can calculate the average bits used for the calibration strategy. 

It should be noted that when both encoding and decoding use single-precision floating point format (i.e., FP32), the error between different platforms is usually small (e.g., the maximum error between the calculations of $\boldsymbol{I}$ using A100 and 2080 is 5e-5). Thus using the calibration strategy proposed for cross-platform decoding will not result in too much additional bitstream.

\begin{figure}[h]
  \centering
  \includegraphics[width=0.5\linewidth]{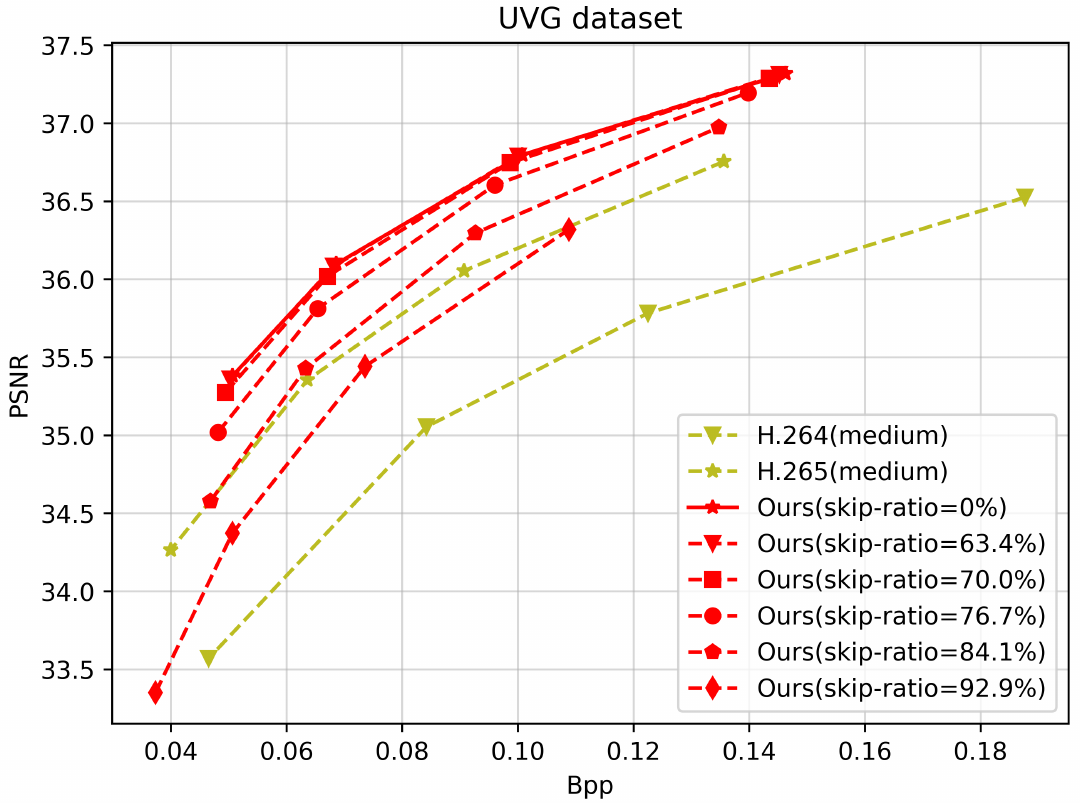}
  \caption{RD curves on UVG dataset with different skip ratios.}
  \label{fig:ab_skip}
\end{figure}

\noindent\textit{\textbf{Skipping of Arithmetic Coding.}}
In addition to strategies such as model pruning and motion downsampling, we also introduce the skipping strategy in arithmetic coding. By filtering the estimated $(\boldsymbol{\mu}, \boldsymbol{\theta})$ of entropy coding, a certain proportion of arithmetic coding can be jumped over to achieve computational acceleration at the encoding and decoding ends. As shown in Fig.~\ref{fig:ab_skip}, different RD curves are obtained by performing skipping operations on our model at different filter levels, where the solid red line represents without skipping and the dashed red line represents skipping at different proportions. As shown in Table. \ref{tab:ab_skip_bd_rate}, we calculate the BD-rate improvement at different skipping ratios using H.265 as the anchor. It can be seen that even if arithmetic coding is skipped by more than 70\% proportion, the model's bitrate saving only declines by 1.1 percent.

Considering that arithmetic coding is not parallel-friendly due to the serialization computation. Nevertheless, it is possible to achieve parallel acceleration by partitioning and then parallelizing the encoding process, which is not adopted in this work.

\begin{table}[t]
\caption{BD-rate calculated by PSNR with different skipping ratios on UVG dataset compared with H.265 (medium).}
\centering
\begin{tabular}{c|rrrrrr}
\hline
\hline Skip ratio & 0\% & 63.4\% & 70.0\% & 76.7\% & 84.1\% & 92.9\% \\
\hline
BD-rate & -25.6\% & -25.6\% & -24.5\% & -19.0\% & -6.2\% & 12.8\% \\
\hline
\end{tabular}
\label{tab:ab_skip_bd_rate}
\end{table}

\section{Conclusion}

The major contribution of this paper is to propose a new solution for the cross-platform codec, which adopts a series of effective methods to ensure the indicators and efficiency of the model. It ultimately achieves a video codec that can be decoded in real-time on user devices and has performance surpassing H.265 (medium).

Existing methods for solving cross-platform issues all use quantization strategies, which require more or less training and alignment. Our method can achieve cross-platform consistency without training, just by adding a small amount of calibration information in the transmitted bytes. Theoretically, it is effective for all encoding and decoding methods that use entropy models and arithmetic coding for actual transmitting, which is conducive to the acceleration and implementation of existing compression methods. 

However, transmitting calibration information is absolutely related to cross-platform computational errors. When there is a huge error in cross-platform calculation (e.g., 1e-1), the bitstream of calibration information to be transmitted will become too large, which will decrease the rate-distortion performance.

In summary, significant research progress has been made in neural video codecs. The future prospects for practical applications are immense with more work pushing forward.

\bibliographystyle{unsrt}  
\bibliography{references}

\end{document}